\documentclass[letterpaper]{article}
\usepackage{accelerated_discovery_of_sustainable_building_materials}

\usepackage{graphicx}
\usepackage{subcaption}
\usepackage{times}
\usepackage{helvet}
\usepackage{courier}
\usepackage{commath}
\usepackage{natbib}
\usepackage{array}
\usepackage{booktabs} 
\usepackage{chemformula}
\usepackage{breqn}
\newcolumntype{P}[1]{>{\centering\arraybackslash}p{#1}}
\newcommand{\KLD}[2]{D_{\mathrm{KL}} (  #1 \left| \right| #2 )}

\title{Accelerated Discovery of Sustainable Building Materials
}

\author{
  Xiou Ge\\
  University of Illinois at Urbana-Champaign
  \And
  Richard T. Goodwin\\
  IBM T.J. Watson Research Center
  \And
  Jeremy R. Gregory\\
  Massachusetts Institute of Technology
  \AND
  Randolph E. Kirchain\\
  Massachusetts Institute of Technology
  \And
  Joana Maria\\
  IBM T.J. Watson Research Center
  \And
  Lav R. Varshney\\
  University of Illinois at Urbana-Champaign
}
\begin{document} 
\maketitle
\begin{abstract}
\begin{quote}

Concrete is the most widely used engineered material in the world with more than 10 billion tons produced annually. Unfortunately, with that scale comes a significant burden in terms of energy, water, and release of greenhouse gases and other pollutants. As such, there is interest in creating concrete formulas that minimize this environmental burden, while satisfying engineering performance requirements.  Recent advances in artificial intelligence have enabled machines to generate highly plausible artifacts, such as images of realistic looking faces. Semi-supervised generative models allow generation of artifacts with specific, desired characteristics. In this work, we use Conditional Variational Autoencoders (CVAE), a type of semi-supervised generative model, to discover concrete formulas with desired properties. Our model is trained using open data from the UCI Machine Learning Repository joined with environmental impact data computed using a web-based tool. We demonstrate CVAEs can design concrete formulas with lower emissions and natural resource usage while meeting design requirements. To ensure fair comparison between extant and generated formulas, we also train regression models to predict the environmental impacts and strength of discovered formulas. With these results, a construction engineer may create a formula that meets structural needs and best addresses local environmental concerns.

\end{quote}
\end{abstract}

\section{Introduction}

The building sector accounts for a significant proportion of overall energy consumption and pollution worldwide. Concrete, including its primary ingredient, cement, is one of the most energy intensive and polluting building materials to fabricate. Building environmental-friendly infrastructure and reducing pollution due to rapid urbanization are two of the Sustainable Development Goals (SDGs) to be achieved by 2030. Given a desire for more sustainable development, there is growing interest in discovering concrete formulas that minimize pollution. As an example, using low carbon footprint concrete will lead to improvements in Indicator 9.4.1 of the SDG,\footnote{https://sustainabledevelopment.un.org/sdg9} which evaluates the world's progress to the SDG target by measuring $\ch{CO2}$ emission from unit economic activities.

In the automated material discovery domain, the Materials Project,\footnote{https://www.materialsproject.org/} a main program of the Materials Genome Initiative \citep{Jain2013}, is a web-based platform that provides both open source data sets and data analysis tools for researchers to design novel materials. The tool has a relatively large and comprehensive database and interactive tools for materials such as inorganic compounds, nonporous materials, electrodes, etc. However, the Materials Project has not extended to the concrete mixture design domain which we think is equally important.

To this end, the Cement Sustainability Initiative (CSI) developed the Environmental Product Declaration (EPD) tool to facilitate the generation of sector-specific EPDs for cement and concrete but also for clinker, lime, and plaster.\footnote{http://wbcsdcement.org/epd-tool-1} EPD is a voluntary declaration that provides quantitative information about the environmental impact of a product, using life-cycle assessment (LCA) methodology and verified by an independent third party. The cloud-based tool was designed to be easy-to-use, to facilitate the process overall, and to reduce the costs of preparing cement and concrete EPDs. In this work, we join this with data from the open UCI repository.

Furthermore, recent advances in artificial intelligence have enabled machines to generate highly plausible artifacts, such as images of realistic looking faces \citep{Attribute2Image}, and natural language \citep{DBLP:journals/corr/BowmanVVDJB15}. In this work, we use Conditional Variational Autoencoders (CVAE), a type of semi-supervised generative model, to generate concrete formulas with desired properties. We demonstrate CVAEs can design concrete formulas with lower emissions and natural resource usage while meeting design requirements. To ensure fair comparison between extant and generated formulas, we also train regression models to predict the environmental impacts and strength of generated formulas. With these results, a construction engineer may create a formula that meets structural needs and best addresses local environmental concerns.

The rest of this paper is organized as follows. We first survey some past work on applying data science to scientific discovery. We then move on to describe the data set and the CVAE model details. Next we give results, first showing the average percentage reduction environmental impact achieved by generated better-performing concrete formulas. We then show strength spectrum plots in the 3D environmental impact space which could be turned into a visualization tool for concrete designers. Lastly, we evaluate the performance of strength conditioned generation of the trained model. 

\section{Related Work}

Data science has been applied to scientific discovery for some time but is now gaining popularity. Within materials discovery, the Discovery through Eigenbasis Modeling of Uninteresting Data (DEMUD) algorithm proposed by \citet{Wagstaff:2013:GSD:2891460.2891586} guides scientific discovery by prioritizing the data point that carries more information to investigate and backing up with an explanation for novelty, using dimensionality reduction techniques. \citet{Varshney2018} proposes to use Bayesian surprise \citep{NIPS2005_2822} as an objective to select the most interesting data point for investigation. \citet{Balachandran2016AdaptiveSF} use a regressor-selector pair to locate the desired material in the least number of iterations by alternating between exploration and exploitation strategies. 

Recently, deep generative models have been applied in materials and molecules discovery. \citet{vae_molecule} use variational autoencoders (VAEs) based on recurrent neural networks (RNNs) for chemical design in which molecules are encoded as strings. \citet{2017arXiv170608203R} use VAEs to improve the accuracy of drug response predictions. Moreover, semi-supervised generative models allow generation of artifacts with specific, desired characteristics \citep{Attribute2Image}. We transfer this idea to the concrete formulation generation task to make the synthesis controllable.

Comprehensive physics-based models that predict concrete performance from formulas have been elusive for a century \citep{mehta1986concrete}. Recently, discriminative machine learning models have been applied to predict the compressive strength of concrete and demonstrated good performance \citep{CHOU2014771}. However, the environmental impacts of concrete have never been considered in terms of predictive machine learning models. Moreover, the success of deep generative models in generating realistic visual and audio data have given hope to generate other artifacts such as concrete formulas. Here we develop a novel generative machine learning model---a form of computational creativity or accelerated discovery \citep{Besold2015ComputationalCR}---with the capability to design environment-friendly concrete that may help in meeting sustainable development targets. 

\section{Approach}

\begin{figure}
\centering
\includegraphics[width=\columnwidth]{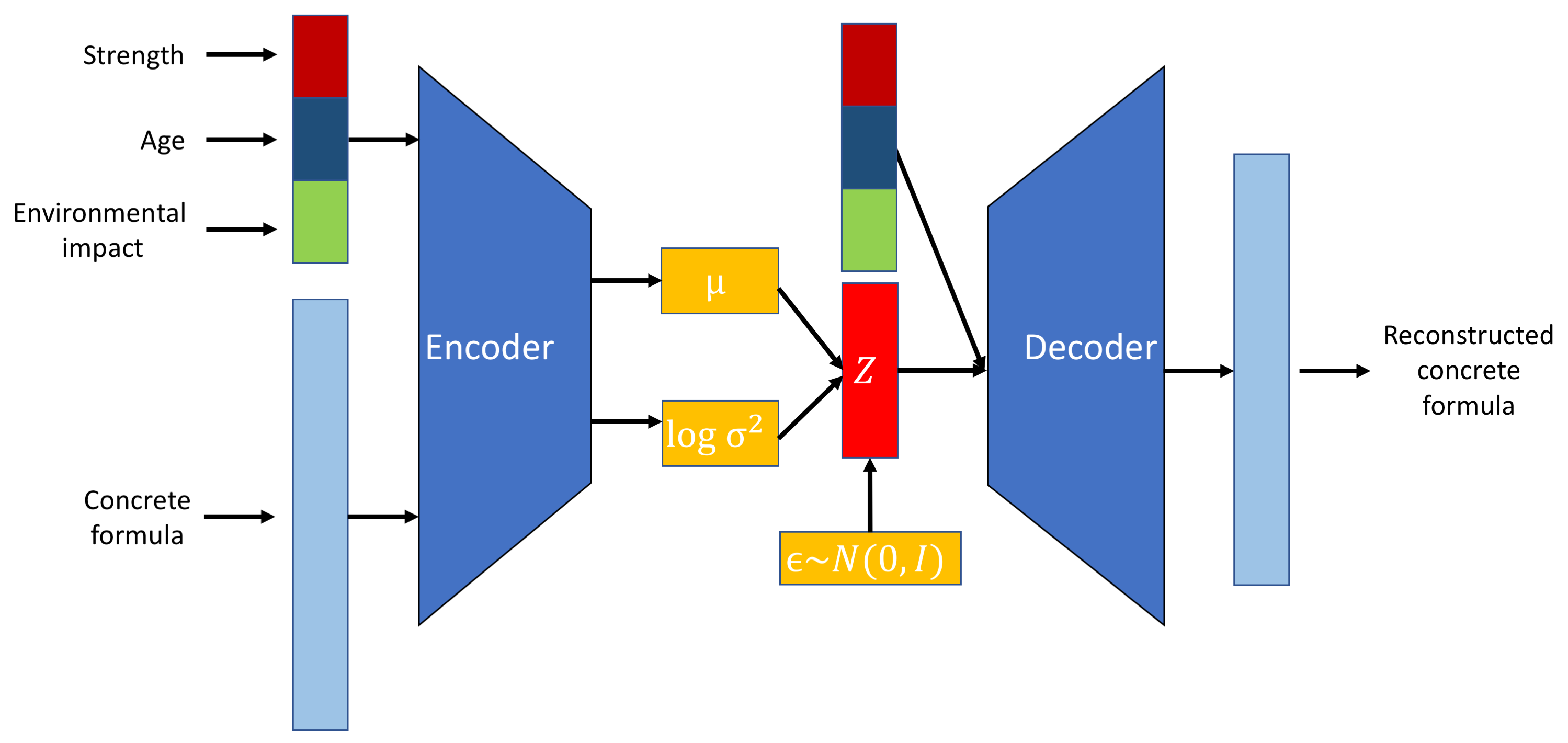}
\caption{CVAE Model Structure}
\label{fig:CVAE_structure}
\end{figure}

\subsection{Data set}
We train our model using the Concrete Compressive Strength Data Set \citep{Yeh19981797} openly available from the UCI Machine Learning Repository. It has 1,030 training examples, with seven continuous features describing the amount of constituent material such as cement, aggregates, and water. Compressive strength, after a particular curing time (age), of each concrete formula is also given. In addition, we use the CSI EPD tool to estimate the environmental impact of each concrete formula. The EPD tool produces 12 continuous features characterizing the concrete environmental impact. Among these, we largely focus on global warming potential (GWP), acidification potential (AP), and concrete batching water (CBW) consumption.

\subsection{Generative Model}
Our model is based on a variant of the VAE \citep{2013arXiv1312.6114K} called CVAE \citep{NIPS2015_5775} as shown in Fig.~\ref{fig:CVAE_structure}. Like other generative models, the goal is to estimate the data distribution $p(y)$ and to generate realistic new samples from that distribution \citep{2016arXiv160605908D}. What makes CVAE different from VAE is that instead of merely generating realistic samples from the data distribution $p(y)$ randomly, we generate from the conditional distribution $p(y|x)$ which give us control over the underlying properties of generated data by conditioning on different values of $x$. 

We interpret the variables in the conditional generative model as follows: $x$ represents the side information of a formula including the strength, age, and environmental impacts, $y$ represents the constituent material amount of a formula, and $z$ is the latent variable. Like the VAE, a CVAE consists of an encoder $q_\phi(z|x,y)$ that maps the data points to latent codes and a decoder $p_\theta(y|x,z)$ that reconstructs the data points from latent codes. The decoder and encoder are implemented as neural networks where $\phi$ and $\theta$ are the respective network parameters. Since the goal is to generate realistic concrete samples with desired properties, we want to maximize the log likelihood of the data distribution model $\log{p_\theta(y^{(i)}|x^{(i)})}$. Since the data  distribution $p_\theta(y|x)$ and the posterior distribution $p_\theta(z|x,y)$ are both intractable, we maximize the Evidence Lower Bound (ELBO), $\mathcal{L}$, instead. The loss function of CVAE is:
\begin{multline}
	\log{p_\theta(y^{(i)}|x^{(i)})} \geq \mathbf{E}_{z\sim q_\phi(z|x,y)}[\log{p_\theta(y^{(i)}|z,x^{(i)})}] \\
	- \KLD{q_\phi(z|x^{(i)},y^{(i)})}{p_\theta(z|x^{(i)})} = \mathcal{L}
\end{multline}

\subsection{Implementation Details}

In our model, the encoder network consists of four fully-connected layers with 25 neurons on the first layer, 20 neurons on the second layer, followed by two parallel layers with two neurons on each which represent the mean and log variance respectively. The prior is set to be an isotropic Gaussian distribution with zero mean and unit variance $p(z)=\mathcal{N}(0,\mathcal{I})$. The reparameterization trick is performed to make the sampling step differentiable and enable backpropagation for training. The decoder network consists of two fully-connected layers with 20 neurons on the first layer and 25 neurons on the second layer. ReLU activation functions are applied to all layers except the output layer of the decoder, where we use sigmoid activation since we scale our data to $[0,1]$. The model is trained end to end with Adam optimizer with learning rate of 0.001 and batch size of 10.

\begin{figure}
\centering
\includegraphics[width=\columnwidth]{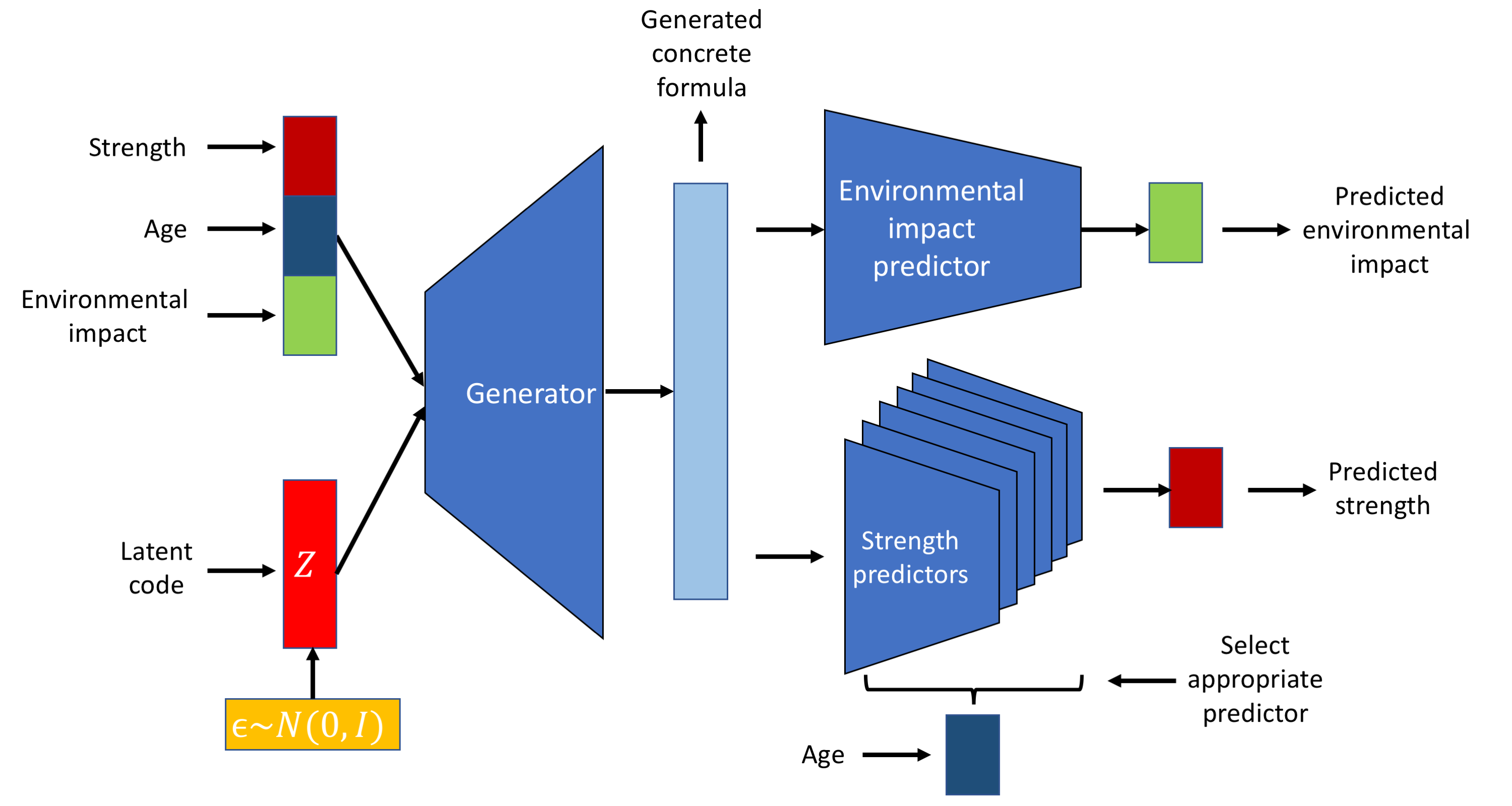}
\caption{Generating new concrete formulas and evaluating their properties}
\label{fig:generator_predictor}
\end{figure}

\subsection{Property Predictors}
We also trained neural network-based regression models as shown in Fig.~\ref{fig:generator_predictor} using the data set that we described above to predict the environmental impact and strength of concrete formulas. Since the compressive strength is dependent on the age of concrete, we trained separate compressive strength predictors for each age group. The purpose of the predictors is twofold. First, we can measure how well the properties of generated samples match the desired properties given as conditioning variables during data generation. Second, we can make fair comparisons between extant and generated concrete formulas in terms of the environmental impact. We experimented with three different types of regression models, namely linear regression, decision tree regression, and neural network regression. Although linear regression can achieve comparable performance with decision tree regression and neural network regression, it often predicts far-out-of-range values for newly generated concrete formulas. The neural network regression has slightly better performance than the decision tree regression and therefore we use the former for prediction tasks. The performance of the neural network regression models for global warming potential (GWP), acidification potential (AP), and concrete batching water (CBW) consumption are shown in Table~\ref{table:Predictor_Performance}. The performance of the strength predictors are shown in Table~\ref{table:Strength_Predictor_Performance}.
 
\begin{table}
\centering
\begin{tabular}{rrrrr}
\toprule
Metric & GWP & AP &CBW \\
 & (kg $\ch{CO2}$ eq.) & (kg $\ch{SO2}$ eq.) & ($m^3$) \\
\midrule
MAE & 7.187  & 0.019 & 0.003   \\
RMSE & 9.374  & 0.040 & 0.006  \\
$R^2$ & 0.979  & 0.974 & 0.881 \\
\bottomrule
\end{tabular}
\caption{Environmental Impacts Predictor Performance}
\label{table:Predictor_Performance}
\end{table}

\begin{table}
\centering
\begin{tabular}{rrrrrrr}
\toprule
& 
& \multicolumn{5}{c}{Predictor Performance (MPa)}\\
\cmidrule(r){2-7}
Metric & $\leq$3 & 7 & 14 & 28 & 56 & $\geq$90 \\
\midrule
MAE & 2.985 & 3.850 & 3.378 & 6.015 & 5.093 & 4.457\\
RMSE & 0.222 & 0.201 & 0.163 & 0.227 & 0.124 & 0.125\\
$R^2$ & 0.819 & 0.870 & 0.703 & 0.679 & 0.795 & 0.789\\
\bottomrule
\end{tabular}
\caption{Strength Predictor Performance}
\label{table:Strength_Predictor_Performance}
\end{table}

\section{Results}

\subsection{Generating environmental impact reducing concrete formulas}

\begin{table}
  \centering
  \begin{tabular}{rrrrr}
    \toprule
     & & \multicolumn{3}{c}{Average Reduction ($\%$)}\\
    \cmidrule(r){3-5}
    Age & Strength &  GWP & AP & CBW \\
    (day) & (MPa) &  (kg $\ch{CO2}$ eq.) &  (kg $\ch{SO2}$ eq.)& ($m^3$) \\
    
    \midrule
    $\leq$3 & 30$\pm$1 & 0.80 & 1.83 & 5.47 \\
            & 40$\pm$1 & 7.74 & 1.59 & 0.26 \\
    \cmidrule(r){1-5}
    7 & 30$\pm$1 & 19.69 & 3.94 & 7.58 \\
      & 40$\pm$1 & 25.45 & 11.33 & 5.03 \\
    \cmidrule(r){1-5}
    14 & 20$\pm$1 & 2.20 & 5.72 & 10.64 \\
       & 60$\pm$1 & 42.45 & 21.09 & 5.17 \\
    \cmidrule(r){1-5}
    28 & 70$\pm$1 & 21.62 & 6.66 & 3.32 \\
       & 80$\pm$1 & 27.44 & 8.40 & 4.15 \\
    \cmidrule(r){1-5}
    56 & 40$\pm$1 & 4.38 & 2.95 & 7.04 \\
       & 50$\pm$1 & 14.38 & 3.23 & 3.64 \\
       & 70$\pm$1 & 30.26 & 23.75 & 1.32 \\
       & 80$\pm$1 & 5.88 & 1.33 & 3.46 \\
    \cmidrule(r){1-5}
    $\geq$90 & 80$\pm$1 & 30.58 & 6.91 & 4.11 \\
    \bottomrule
  \end{tabular}
  \caption{Average environmental impact reduction achieved by better performing generated samples}
\label{table:environmental_impact_reduction}
\end{table}

\begin{table}[t]
  \centering
  \begin{tabular}{rrr}
    \toprule
    Strength (MPa) & 30$\pm$1 & 40$\pm$1 \\
    \cmidrule(r){2-3}
    Constituent Material & \multicolumn{2}{c}{Amount (kg per $m^3$)}\\
    \midrule
    Cement & 186.4 & 259.0\\
    Blast Furnace Slag & 236.7 & 288.6\\
    Fly Ash & 107.1 & 58.8\\
    Water & 142.3 & 142.5\\
    Superplasticizer & 22.3 & 26.1\\
    Coarse Aggregate & 901.4 & 868.6\\
    Fine Aggregate & 717.2 & 763.0\\
    \bottomrule
  \end{tabular}
  \caption{Sample concrete formula with reduced environmental impact}
  \label{table:sample}
\end{table}

\begin{figure*}[htb]
    \centering 
\begin{subfigure}{0.5\textwidth}
  \includegraphics[width=\linewidth]{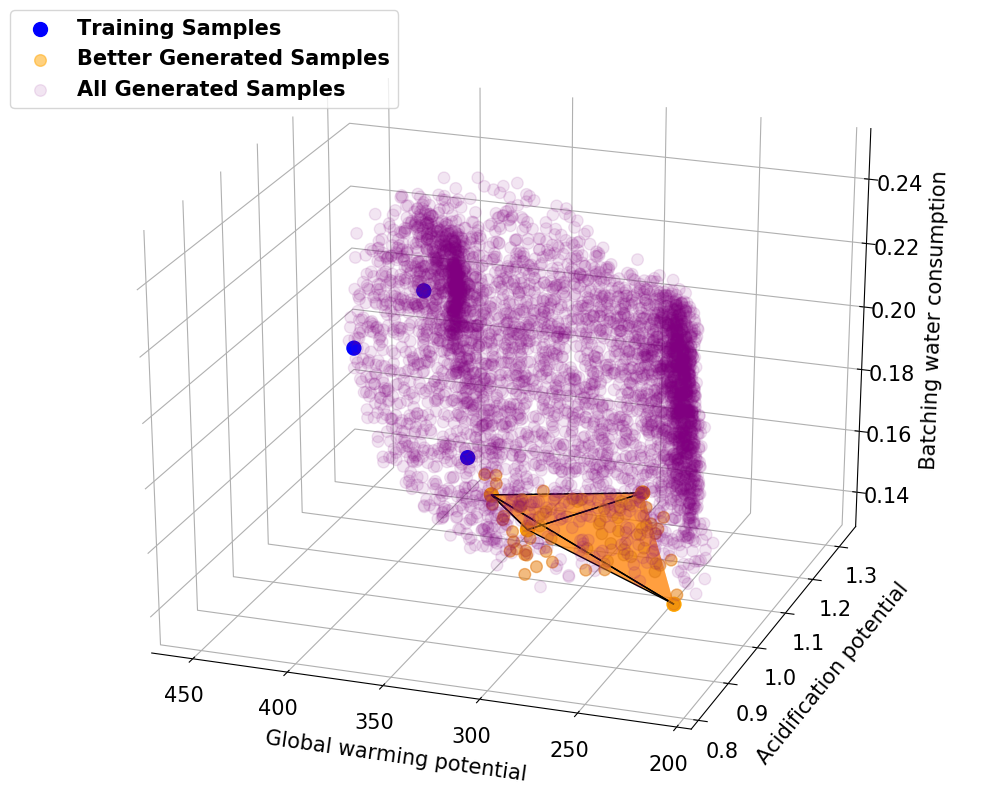}
  \caption{Curing time = 7 days, Strength = 30$\pm$1 MPa}
  \label{fig:day_7_strength_30}
\end{subfigure}\hfil 
\begin{subfigure}{0.5\textwidth}
  \includegraphics[width=\linewidth]{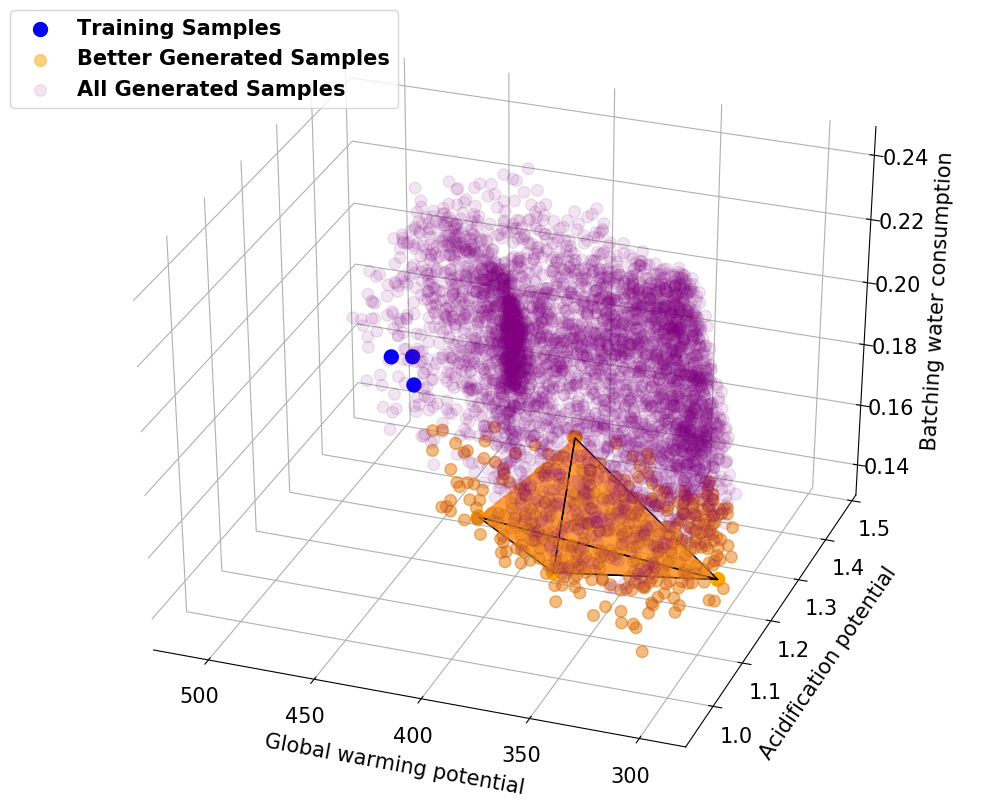}
  \caption{Curing time = 7 days, Strength = 40$\pm$1 MPa}
  \label{fig:day_7_strength_40}
\end{subfigure}
\caption{Approximated hull of generated samples from archetypal analysis, training samples, and all generated samples for specific curing time and strength level.}
\label{fig:convex_hull}
\end{figure*}

To demonstrate that the generative algorithm discovers new concrete formulas with reduced environmental impacts, we compared the GWP, AP, and CBW values between extant concrete formulas and generated formulas with the same age and similar strength. For each concrete age group, we generate 60,000 concrete formulas. Both the strength and the environmental impact inputs to the generator are produced by randomly sampling from the standard uniform distribution whereas the latent code input is produced by randomly sampling from the standard bivariate normal distribution. We then use the trained environmental impact predictor and strength predictor for the corresponding age group to evaluate environmental impact and strength of the newly generated formulas. We count the number of generated samples having lower environmental impact than the best observed values for extant samples in all 3 dimensions. We also measured the average percentage reduction in environmental impact for the better-performing samples as compared to extant samples. 

Results shown in Table~\ref{table:environmental_impact_reduction} indicate that there is significant opportunity to reduce environmental impact. We constructed an approximated convex hull that encloses a majority of the better performing points in the 3D space as shown in Fig.~\ref{fig:convex_hull}. From the diagram we can also see that there is an opportunity to trade off different types of environmental impact. In Table~\ref{table:sample}, we show one specific generated concrete formula that is nearest neighbor to one of the extremal points used to construct the convex hull, for strength of 30$\pm$1 MPa and 40$\pm$1 MPa respectively. 

\subsection{Visualization for concrete designers}

\begin{figure*}[htb]
    \centering 
\begin{subfigure}{0.33\textwidth}
  \includegraphics[width=\linewidth]{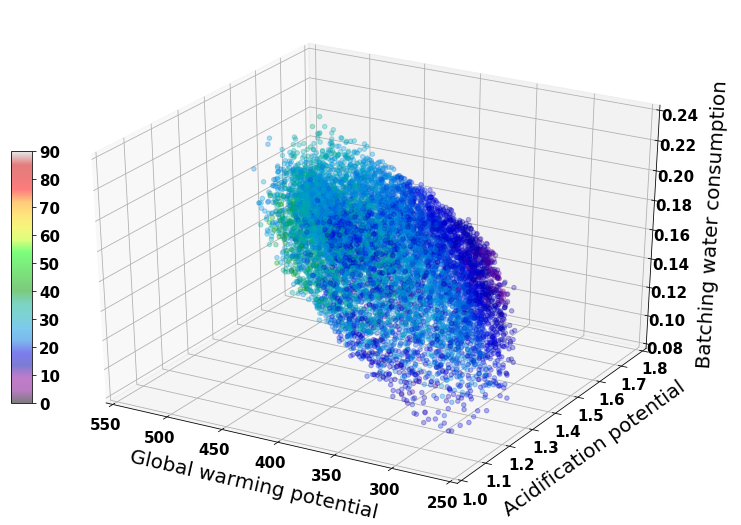}
  \caption{$\leq$ 3 days}
  \label{fig:strength_env_profile_0}
\end{subfigure}\hfil 
\begin{subfigure}{0.33\textwidth}
  \includegraphics[width=\linewidth]{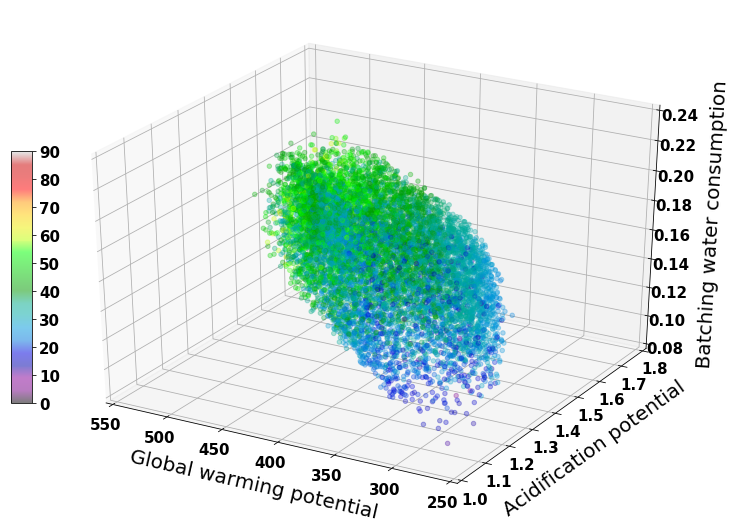}
  \caption{7 days}
  \label{fig:strength_env_profile_1}
\end{subfigure}\hfil
\begin{subfigure}{0.33\textwidth}
  \includegraphics[width=\linewidth]{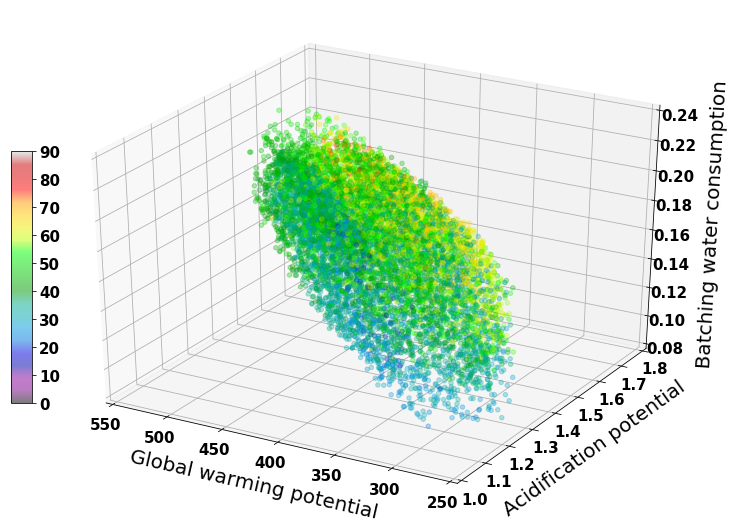}
  \caption{14 days}
  \label{fig:strength_env_profile_2}
\end{subfigure}

\smallskip
\begin{subfigure}{0.33\textwidth}
  \includegraphics[width=\linewidth]{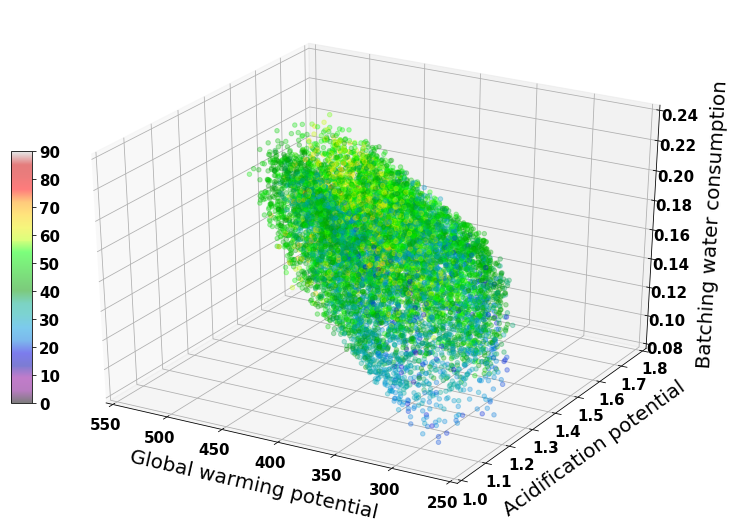}
  \caption{28 days}
  \label{fig:strength_env_profile_3}
\end{subfigure}\hfil 
\begin{subfigure}{0.33\textwidth}
  \includegraphics[width=\linewidth]{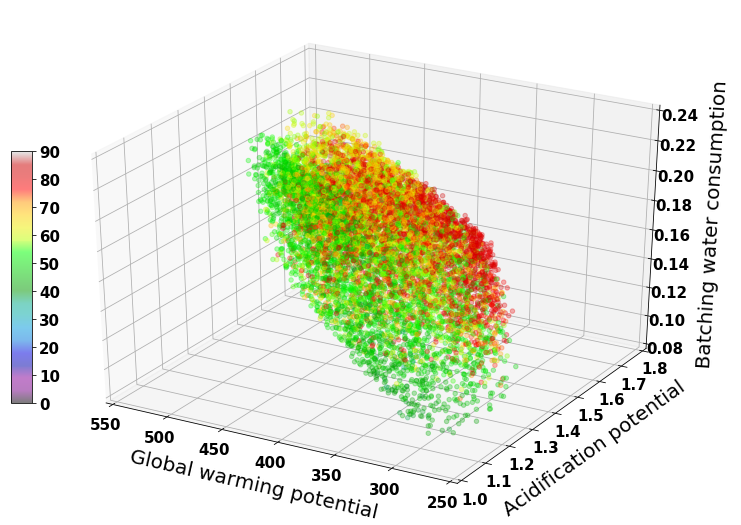}
  \caption{56 days}
  \label{fig:strength_env_profile_4}
\end{subfigure}\hfil
\begin{subfigure}{0.33\textwidth}
  \includegraphics[width=\linewidth]{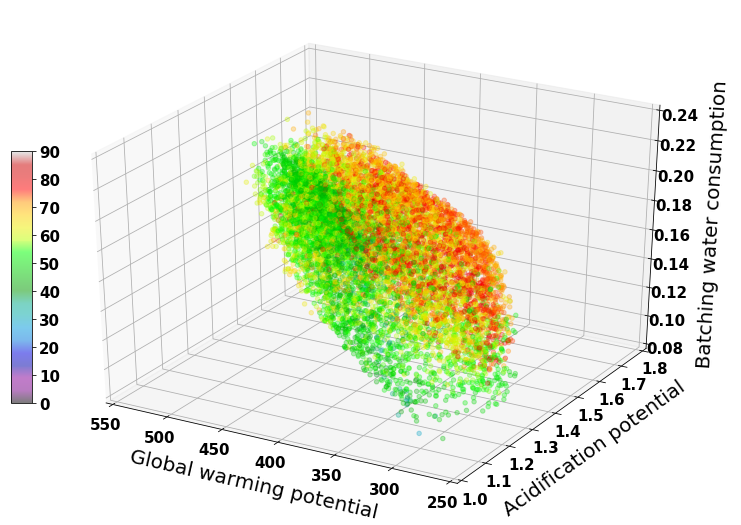}
  \caption{$\geq$90 days}
  \label{fig:strength_env_profile_5}
\end{subfigure}
\caption{Strength spectrum of generated concrete formulas for different concrete curing time plotted in 3D environmental impacts space, where color indicates strength}
\label{fig:strength_env_profile}
\end{figure*}

On top of the 3D environmental impact design space that we mentioned earlier, we also color each data point based on the predicted strength of the corresponding formula. Fig.~\ref{fig:strength_env_profile} shows the strength spectrum of the newly generated concrete formulas plotted in the environmental impact space for each concrete curing time group. These plots could serve as a visualization tool for the concrete designers to quickly select newly generated formulas that meet the design requirements.

\subsection{Strength-conditioned progression generation}

\begin{figure*}[htb]
    \centering 
\begin{subfigure}{0.33\textwidth}
  \includegraphics[width=\linewidth]{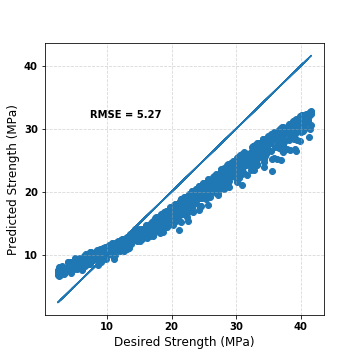}
  \caption{$\leq$ 3 days}
  \label{fig:strength_progression_0}
\end{subfigure}\hfil 
\begin{subfigure}{0.33\textwidth}
  \includegraphics[width=\linewidth]{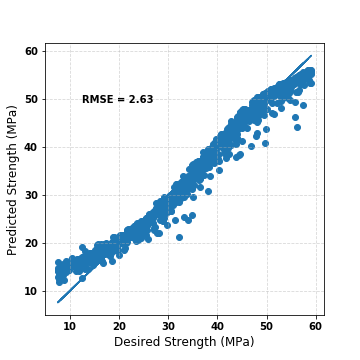}
  \caption{7 days}
  \label{fig:strength_progression_1}
\end{subfigure}\hfil 
\begin{subfigure}{0.33\textwidth}
  \includegraphics[width=\linewidth]{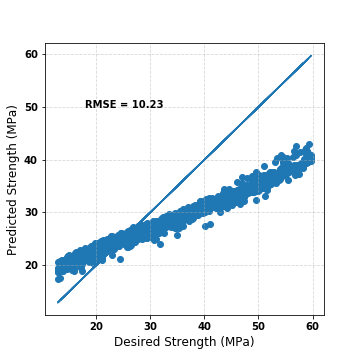}
  \caption{14 days}
  \label{fig:strength_progression_2}
\end{subfigure}

\medskip
\begin{subfigure}{0.33\textwidth}
  \includegraphics[width=\linewidth]{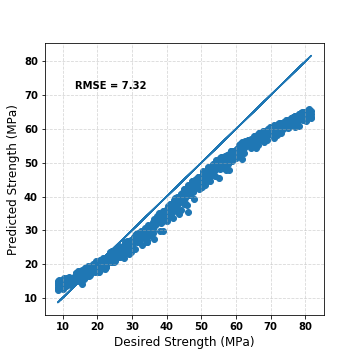}
  \caption{28 days}
  \label{fig:strength_progression_3}
\end{subfigure}\hfil 
\begin{subfigure}{0.33\textwidth}
  \includegraphics[width=\linewidth]{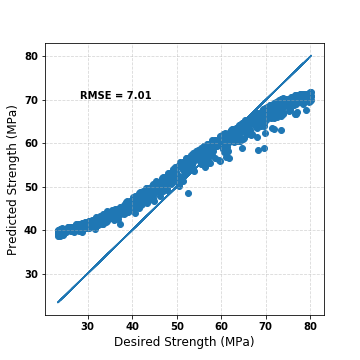}
  \caption{56 days}
  \label{fig:strength_progression_4}
\end{subfigure}\hfil 
\begin{subfigure}{0.33\textwidth}
  \includegraphics[width=\linewidth]{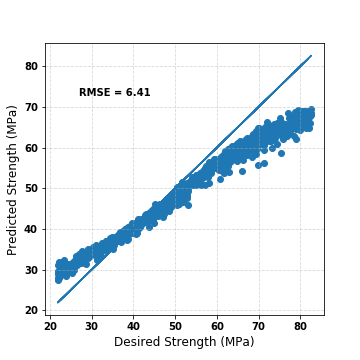}
  \caption{$\geq$90 days}
  \label{fig:strength_progression_5}
\end{subfigure}
\caption{Strength conditioned progression for different concrete curing time}
\label{fig:strength_progression}
\end{figure*}

Attribute-conditioned image progression has been experimented by \citet{NIPS2015_5775}. In the experiment, one of the attribute dimension values such as gender, facial expression, or hair color is modified by interpolating between the minimum and maximum attribute value, i.e. $x = [x_\alpha, x_{rest}]$, where $x_\alpha = (1-\alpha)\cdot x_{min} + \alpha\cdot x_{max}$. Indeed, one can visualize that the attribute of generated images change progressively with the change in conditioning attribute values. 

To further demonstrate our concrete generator can produce concrete designs with desired properties, we perform similar experiments. For the purpose of illustration, we limit our conditioning variables to strength and curing time of the concrete. We again generate 10,000 samples for each curing time group, by uniformly sampling from $[x_{min}, x_{max}]$. Fig.~\ref{fig:strength_progression} shows how well the predicted strength of generated formulations match with the desired strength given as conditioning variable during generation. The performance varies across different curing time groups. RMSE is computed to evaluate the performance quantitatively. The better performing model should have the contour of the scattered dots to cover the diagonal line. The result shows that the generator seems to work the best for concrete curing time of 7 days.

\section{Conclusion}
We have demonstrated that with the data obtained from an open source database and cloud-based tools, we are able to train a CVAE model and discover new concrete formulations with reduced environmental impact. However, there is still room for improving our model. Our data contains both continuous and categorical values, but CVAE may not be the best for capturing such mixed categorical and continuous data. The VAE-ROC model proposed by \citet{2016arXiv160404960S} is said to be better at handling mixed data. We hope by modifying the CVAE model in line with specifics of the VAE-ROC, the generator would synthesize more realistic concrete designs and achieve better performance in attribute-conditioned generation. Experimental verification by actually making newly discovered concrete formulations also remains.

\bibliographystyle{accelerated_discovery_of_sustainable_building_materials}
\bibliography{conf_abrv,abrv,accelerated_discovery_of_sustainable_building_materials}

\end{document}